\documentclass[a4paper]{article}
\usepackage{makecell, booktabs, array}
\usepackage{url}
\usepackage{graphicx}
\usepackage{subfigure}
\usepackage{INTERSPEECH2021}
\title{Arabic Code-Switching Speech Recognition using Monolingual Data}
\name{Ahmed Ali$^1$, Shammur Chowdhury$^1$, Amir Hussein$^{1,2}$,  Yasser Hifny$^3$ }
\address{
  $^1$Qatar Computing Research Institute, HBKU, Doha, Qatar\\
  $^2$Kanari AI, California, USA \\
  $^3$University of Helwan, Egypt}
  
\email{amir@kanari.ai, \{shchowdhury, amali\}@hbku.edu.qa, yhifny@fci.helwan.edu.eg}

\begin{document}

\maketitle
\begin{abstract}
Code-switching in automatic speech recognition (ASR) is an important challenge due to globalization. Recent research in multilingual ASR shows potential improvement over monolingual systems. We study key issues related to multilingual modeling for ASR through a series of large-scale ASR experiments.  
Our innovative framework deploys a multi-graph approach in the weighted finite state transducers (WFST) framework. 
We compare our WFST decoding strategies with a transformer sequence to sequence system trained on the same data. Given a code-switching scenario between Arabic and English languages, our results show that the WFST decoding approaches were more suitable for the intersentential code-switching datasets. In addition, the transformer system performed better for intrasentential code-switching task. With this study, we release an artificially generated development and test sets, along with ecological code-switching test set, to benchmark the ASR performance.

\end{abstract}

\section{Introduction} \label{sec:intro}
Multilingual speech recognition has made rapid progress in recent years, primarily due to the advances in deep learning and the availability of adequate training resources \cite{4960588,6639081,6639348,5495646}.

In a study by \cite{tong2019investigation}, they discussed and compared the connectionist temporal classification (CTC) and the end-to-end lattice-free maximum mutual information (LF-MMI) for multilingual ASR. They illustrated that end-to-end LF-MMI is indeed competitive on a low-resourced multilingual task, comfortably outperforming a CTC baseline, owing specifically to its end-to-end nature, while the usage of context-independent phone labels made it attractive for multilingual ASR.

%A large scale multilingual hybrid ASR system has been studied by \cite{liu2019multilingual}, where they developed a single grapheme-based speech recognition model learned on 7 languages, using standard hybrid BLSTM-HMM acoustic models with LF-MMI objective. Their system is based on the union over each language-specific grapheme set. They found such a multilingual graphemic hybrid speech recognition system can outperform language-independent recognition on all 7 languages. Moreover,  it substantially outperformed each monolingual speech recognition system. Their ASR can decode any of the 7 languages, even though not explicitly given any information about language identity. However, their study did not address the code-switching between the  languages during the search process.

%Recently, end-to-end approaches for ASR have received extensive attention. Popular end-to-end approaches are CTC \cite{graves2006connectionist}, attention-based  \cite{chan2016listen} and RNN-Transducers methods \cite{he2019streaming, rao2017exploring, graves2012sequence}. End-to-end models are well suited for multilingual ASR because they encapsulate acoustic, pronunciation and language models jointly in a single network \cite{toshniwal2018multilingual, miiller2018multilingual, datta2020language, biswas2018multilingual}.

In a study by \cite{toshniwal2018multilingual}, they presented a single sequence-to-sequence ASR model trained on 9 different Indian languages, with very little overlap in their scripts. They took the union of language-specific grapheme sets and trained a grapheme-based sequence-to-sequence model jointly on data from all languages. They achieved 21\% in performance over separately trained systems.
A language-agnostic multilingual ASR system was studied by \cite{datta2020language}, which is the case for multicultural societies where several languages are frequently used together and often rendered with different writing systems. The system transforms all languages to one writing system through a many-to-one transliteration transducer. Thus, similar sounding acoustics were mapped to a single, canonical target sequence of graphemes, effectively separating the modeling and rendering problems. They achieved 10\% relative reduction over a language-dependent system.

Most of the previous studies focused on multilingual scenarios without paying attention to code-switching, where there is an alternation between two or more languages within the same utterance.  Code-switching in spontaneous speech is highly unpredictable and difficult to model \cite{sitaram2019survey}. English-Mandarin has %probably
been studied most extensively \cite{li2013improved, li2013language, vu2012first} in addition to other language pairs such as Frisian-Dutch \cite{yilmaz2016longitudinal}, Hindi-English \cite{ pandey2018phonetically, taneja2019exploiting,sreeram2020exploration} and French-Arabic \cite{amazouz2017addressing}. 

There have been initiatives in building and analysing code-switching in Arabic-English speech \cite{ali2021connecting}. \cite{elfahal2019automatic} studied code-switching in Sudanese Arabic and English through social media applications, 75\% of their corpus was ready by 87 bilingual Arabic native speakers resulting in 2,289 audio files. They achieved 33\% word error rate (WER) using ASR and language identification pipeline. \cite{hamed2020arzen} built the ArzEn corpus, speech corpus for Egyptian Arabic and English code-switching. ArzEn comprised of 12 hours transcribed along with meta-data from linguistic, sociological, and psychological perspectives. Their corpus was recorded by 38 participants. They reported Code-mixing Index (CMI) over the whole corpus of 0.12, and over the code-switching sentences is only 0.17. 
A recent study, in \cite{chowdhury2021onemodel}, proposed a multilingual strategy to model code-switching in Arabic ASR. With the E2E model they reported state-of-the-art results indicating the efficacy of such method for handling both cross-lingual and Arabic dialectal code-switching with little CS data present in QASR dataset \cite{qasr}.

However, most of the previous studies in Arabic English code-switching are lacking the following: (\textit{i}) publicly available corpus to reproduce their results; (\textit{ii}) real code-switching (ecological) data where participants are talking spontaneously in daily life, not in data collection projects; and mostly (\textit{iii}) detailed and complete analysis of ASR behaviours and insights on AM and LM system in code-switching ASR. 
To the best of our knowledge, there are no public studies with reproducible results for Arabic-English code-switching in spontaneous speech using the available monolingual dataset.

\noindent Therefore, our contributions in this paper are: (\textit{i}) Using global multilingual WFST approach (our baseline), (\textit{ii}) Developing a novel approach combining the Kleene closure with multi-graph WFST to support multilingual and code-switching for hybrid ASR, and (\textit{iii}) Comparing the two hybrid approaches with end-to-end transformer ASR in code-switching scenarios.

\iffalse 
\begin{itemize}
\setlength\itemsep{0em}
    \item Benchmark ASR results for inter- and intra-sentential Arabic-English code-switching, in \textit{spontaneous speech}, using 3 strategies:
    \begin{itemize}
    \setlength\itemsep{0em}
        \item Using global multilingual WFST approach (our baseline). 
        \item Developing a novel approach combining the Kleene closure with multi-graph WFST to support multilingual and code-switching for hybrid ASR.
        \item Comparing the two hybrid approaches with end-to-end transformer ASR in code-switching scenarios.
    \end{itemize}
    \item Build and release the first multi-dialectal Arabic-English development and test speech corpus for studying both inter- and intra-sentential code-switching.
\end{itemize}
\fi

%This is the first study to compare different strategies for building decoding graphs in hybrid systems; in addition to its comparison with end-to-end transformers. These comparisons are achieved for two different types of code mixing style. Most importantly, we are making the test/dev dataset used in the study available to the research community for further exploration and use the reported results as a  benchmark for \textit{multi-dialectal} Arabic/English inter- and intra-sentinental code switching phenomena in our multi-culture society.
%Without the loss of generality, we demonstrate our ideas using two languages only: English and Arabic. Future work may explore multiple languages.

\section{ASR Architecture} \label{sec:asr}
The hybrid HMM-DNN ASR architecture based on the weighted finite-state transducers (WFSTs) outlined in~\cite{mohri2009weighted}. %We describe 
%The monolingual ASR system and its extension to multilingual training. 
The training, development, and testing are the same as the Arabic MGB-2~\cite{ali2016mgb} and the English TED-LIUM3~\cite{hernandez2018ted}. For more details, refer to Table \ref{tab:am_data}.

\iffalse
\subsection{Monolingual ASR} \label{subsec:mono_lingual}
We deploy a grapheme-based acoustic model for the monolingual ASR. A context-dependent acoustic model is constructed using the decision tree clustering of tri-grapheme states, in the same fashion as the context-dependent triphone state tying \cite{young1994tree}.  The recognition experiments are performed using the Kaldi ASR toolkit \cite{povey2011kaldi}. We train a conventional context-dependent Gaussian mixture model-hidden Markov model (GMM-HMM) system with 40k Gaussians using 39-dimensional Mel frequency cepstral coefficient (MFCC) features, including the deltas and delta-deltas to obtain the alignments. These alignments are used for training a Time Delay Neural Network (TDNN) \cite{peddinti2015time} using sequence discriminative training with the LF-MMI objective \cite{povey2016purely}.  The input to the TDNN is composed of 40-dimensional high-resolution MFCC extracted from frames of 25 msec length and 10 msec shift along with 100-dimensional i-vectors computed from 1500 msec. Five consecutive MFCC vectors and the chunk i-vector are concatenated, forming a 300-dimensional features vector each frame. 

A monolingual \textit{4}-gram language model is learned over the transcription of the desired language. %To train the language model we used ....n=? }
\fi

\begin{table}[t]
\centering 
\scalebox{0.75}{
\begin{tabular}{c|c|c|c} 
Type & Hours & Programs & \#segments\\ 
\hline
 AR Train & 1,200h & 2,214 & 370K \\
 AR Dev & 10h & 17 & 5,800 \\
 AR Test &  10h & 17 & 5,600 \\
\hline
 EN Train  & 450h & 2,351 & 268K \\
 EN Dev & 1.6h & 8 & 507 \\
 EN Test &  2.6h & 11 & 1,150 \\
\hline
 \end{tabular}
}
\caption{\textit{Data used for acoustic modeling.}} % training, development and evaluation.}}
\label{tab:am_data}
\vspace{-1.0cm}
\end{table}

%languages, a multilingual graphemic set is obtained by taking a union of each grapheme set from each language considered, each of which can be either overlapping or non-overlapping.

%\subsection{Multilingual Hybrid ASR} \label{subsec:multi_asr}
\noindent \textbf{Multilingual Hybrid ASR:} 
For the hybrid ASR, we trained a Time Delay Neural Network (TDNN) \cite{peddinti2015time} using sequence discriminative training with the LF-MMI objective \cite{povey2016purely} with the alignments from a context-dependent Gaussian mixture model-hidden Markov model (GMM-HMM). The input to the TDNN is composed of 40-dimensional high-resolution MFCC extracted from 25 msec frames and 10 msec shift along with 100-dimensional i-vectors computed from 1500 msec. Five consecutive MFCC vectors and the chunk i-vector are concatenated, forming a 300-dimensional features vector each frame.

A universal grapheme set has been investigated to build end-to-end multilingual ASR systems \cite{ kim2018towards, toshniwal2018multilingual}. Modeling graphemes implicitly models spelling, which reduced the amount of entries in the lexicon. However, graphemes can differ immensely from language to language, and languages may have nothing in common in terms of graphemes  (e.g. Arabic and English in our case). Thus, we propose a multilingual architecture that merges all graphemes from multiple languages, keeping the language identity at the grapheme level.
A multilingual \textit{4}-gram language model is learned over the transcription for all the languages. %, using a similar approach to the LM of monolingual ASR.
%Intuitively, the  number of leaves for the proposed multilingual model tends to be larger than a monolingual neural network, which is one of the hyper-parameters to tune in this architecture.

\iffalse
\begin{table*}[!ht]
\centering
%\resizebox{1\textwidth}{!}{
\scalebox{0.85}{
\begin{tabular}{c|c|c}
%\hline
 {\textbf{}}& \textbf{AM Hyperparameters} & \textbf{LM Hyperparameters}\\\hline
 Optimizer/learning rate & Noam / 5 & Adam / 0.0001
  \\ \hline
  Input & batch-bins: $22$M & batch-size: $64$
  \\ \hline
 Encoder& $12$ layers, $8$ attention heads$/$layer & $12$ layers, $4$ attention heads$/$layer
\\ \hline
Decoder & $6$ layers, $8$ attention heads/layer & $12$ layers, $4$ attention heads/layer

\\ \hline
$d^{att}$ & $512$ & $512$

\\ \hline

FF & $2,048$ & $2,048$\\ 
\hline
\end{tabular}}
\caption{\textit{Values of the tuned hyperparameters for end-to-end AM and LM transformers obtained using grid search.}}
\label{hyp_ASR_trans}
\end{table*}
\fi

%\subsection{End to End ASR Architecture}\label{sec:S2Sasr}
\noindent \textbf{End to End ASR Architecture:} Our end-to-end ASR system is based on the transformer architecture~\cite{vaswani2017attention}, which consists of two sub-networks: an encoder model and a decoder model. The encoder transforms the input filter bank speech features $\mathbf{X}$ to a latent representation $\mathbf{H}$. Given the $\mathbf{H}$ and previous predicted tokens outputs $Y_{1:i-1}$, the decoder generates the next token $Y_{i}$. %\subsection{Encoder}
Th \textbf{Encoder} is a multi-blocked architecture, where each block consists of a multihead self-attention (MA) module and  positionwise feed-forward (FF) module. The \textbf{Decoder} is similar to the encoder. However, it has an additional  masked self-attention layer. The masked self-attention enforces the  decoder to attend only to earlier positions in the output sequence.

During the training, the transformer ASR predicts the target sequence $\mathbf{Y}$ of tokens from acoustic features $\mathbf{X}$. For text tokenization, the word-piece byte-pair-encoding (BPE)~\cite{kudo2018sentencepiece} is used with BPE size of 5000. The total loss function $\mathcal{L}_{asr}$ is a multi-task learning objective that combines the decoder cross entropy (CE) loss   $\mathcal{L}_{\mathrm{ce}}$ and the CTC loss \cite{graves2006connectionist} $\mathcal{L}_{\mathrm{ctc}}$.

\iffalse
\begin{equation}\label{eq4}
\mathcal{L}_{asr}=\alpha \mathcal{L}_{\mathrm{ctc}}+(1-\alpha) \mathcal{L}_{\mathrm{ce}}
\end{equation}
where $\alpha$ is a weighting factor which was set to $0.3$ in this study. 
We first augment the speech data with the speed perturbation \cite{ko2015audio}. Then we extract 83-dimensional feature frames that include 80-dimensional log mel-spectrogram and pitch features \cite{ghahremani2014pitch} and apply cepstral mean and variance normalization (CMVN).
The acoustic feature frames $\mathbf{X}$ are then sub-sampled with two convolutional
layer with 256 channels, stride size 2
and kernel size 3 as in \cite{dong2018speech}. Furthermore, the resulted mel-spectrogram features were augmented with specaugment approach \cite{park2019specaugment}. During the inference, the acoustic transformer predictions are rescored with transformer language model (LM) via shallow fusion \cite{hori2018end}. The LM model was trained on the transcription text from the speech data only. We use the ESPNET~\cite{watanabe2018espnet} to train the AM and LM transformer models for 50 epochs. Each model was trained  using 4 NVIDIA Tesla V100 GPUs, each with $16$ GB memory, which took around two weeks. Table \ref{hyp_ASR_trans} summarizes the hyper-parameters used for AM and LM transformer architectures.

\fi

\begin{figure*}[ht]
\centering
% \begin{minipage}{1.0\textwidth}
\centering
\scalebox{0.7}{
\includegraphics[width=5.6 in]{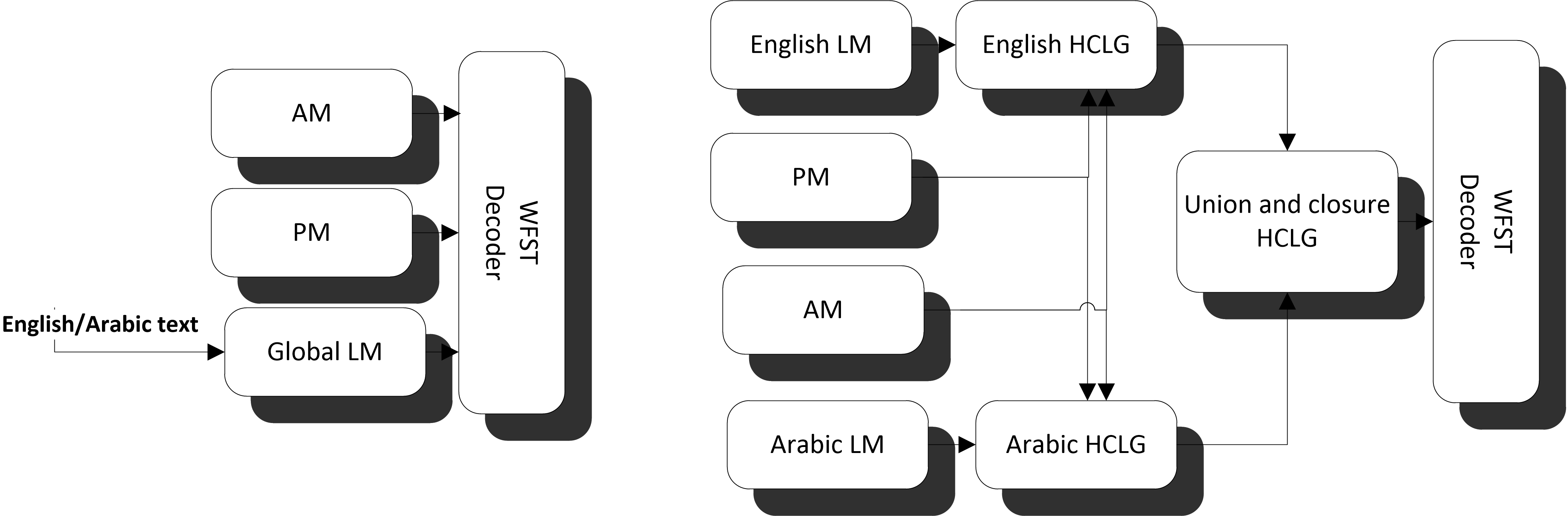}
}
  \caption{\textit{Two approaches to build graphs used for multilingual speech recognition decoding. Global G approach on the left and multi-graph approach on the right. The acoustic model (AM) and the pronunciation model (PM) are shared between the two approaches.}}\label{fig:multi}
% \end{minipage}
\end{figure*}

\section{Multilingual WFSTs}\label{sec:WFST}

A weighted finite-state transducer (WFST) is a generalization of the  finite automata where each transition has an input label, an output label, and an optional weight~\cite{mohri2009weighted}. In this framework, it is possible to combine and compose multiple sources of knowledge to construct the output graph in a unified way. The output graph or network can be optimized using the determination and minimization algorithms. 
The speech recognition decoders based on WFSTs are employed in modern systems \cite{mohri2008speech}. In WFST based decoding, the decoder is used to search the optimal solution constrained by a big WFST. This big WFST graph is composed of a language model (LM), an acoustic model (AM), and lexicon FSTs. Our implementation is based on the Kaldi speech recognition engine~\cite{povey2011kaldi}, where the final decoding graph $HCLG$ is composed of
\begin{equation}
HCLG=H \circ C \circ L \circ G 
\end{equation}
where $H$ is the HMM definitions FST (its input symbols are acoustic modeling
units (transition-ids) and the  output  symbols are the context-dependent phones).
$C$ is the context dependency FST ( its
input symbols represent context-dependent phones and its output symbols are
monophones). $L$ is the pronunciation lexicon FST (its input symbols
are monophones and its output symbols are words). $G$ is the language model or the grammar finite state acceptor (FSA). The HMM transition probabilities, HMM emission probabilities, and LM probabilities are encoded in the graph via the weights. The optimal path is found by searching this graph using the Viterbi algorithm~\cite{forney1973viterbi}. It represents the most likely word sequence given the acoustic features extracted from the input audio.
 
%In a conventional multilingual setup, the recognition is done in two steps. The first step is the language identification from the acoustic features of the input audio. The second step is to decode the input audio based on a graph built for each language as described above. Hence, it is possible to recognize $N$ languages based on $N$ monolingual WFSTs and a language identification system. 
In this work, we present two approaches to build true multilingual speech recognition systems based on one graph. This graph will encode all the knowledge sources for all languages. We assume that the acoustic model and pronunciation models are shared between all languages in this setup. Figure \ref{fig:multi} summarizes the two approaches developed in this work.

\subsection{Global G approach}
In this approach, a language model is trained on a multilingual corpus~\cite{liu2019multilingual}. The corpus consists of the concatenation of the  text for each language. This approach does not require  text  that contains code-switching  where the text is a mix of different languages. The language model is converted into $G$ FSA and composed with other sources to build the $HCLG$ cascaded graph. The graph is used for decoding, and the output of this process is a bilingual/multilingual text, depending on the spoken audio. 

\subsection{Multi-graph approach}

Assume we built a $HCLG$ graph for each language, then it is possible to search the graphs in parallel using a union operation~\cite{yilmaz2019multi}. However, searching the languages in parallel during decoding does not allow the transition between languages. 

To overcome this problem, we propose to add transition arcs between the languages using the Kleene closure. Hence, the decoder can switch between languages during the decoding. % process.%, and  the output of this process is a bilingual/multilingual text, depending on the spoken audio.  

\begin{figure*}[t]
\centering
\begin{subfigure}
    \centering
    \begin{minipage}{0.4\textwidth}
        \centering
        \includegraphics[width=1\textwidth]{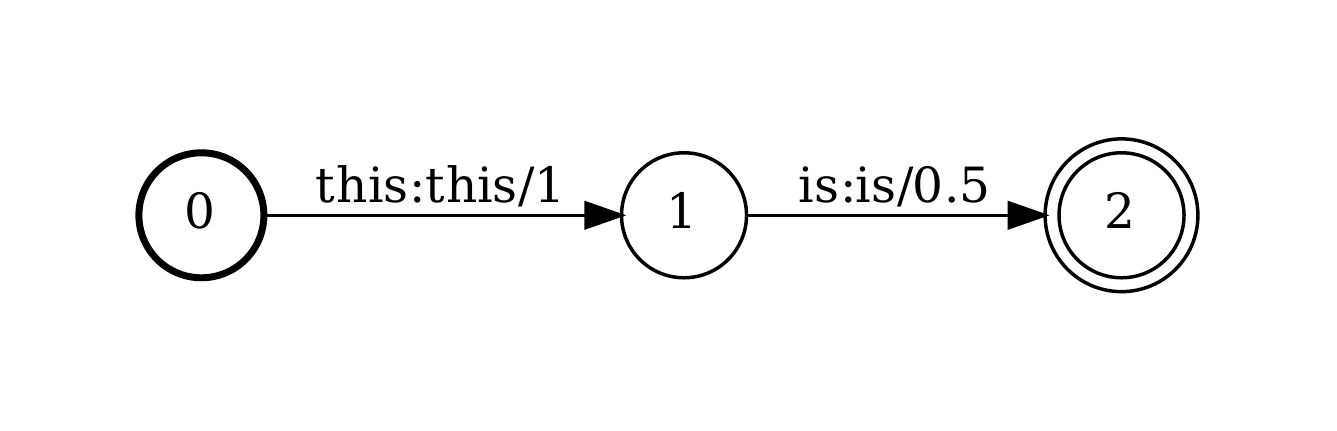}
        %\caption{}
        \label{ch:phase1_cm}
        {\small\textit{(a) A simple $HCLG$ graph for English language.} }
    \end{minipage}\hfill
    \begin{minipage}{0.4\textwidth}
        \centering
        \includegraphics[width=1\textwidth]{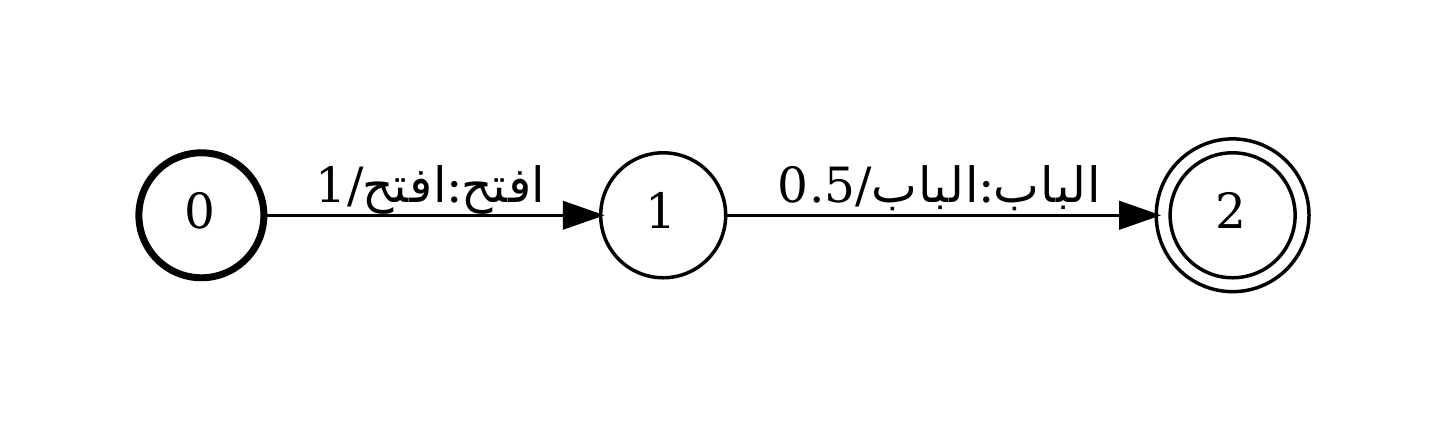}
        %\caption{}
        \label{ch:phase2_cm}
        {\small\textit{(b) A simple $HCLG$ graph for Arabic language.}}
    \end{minipage}
     \caption{\textit{Simple graphs for English and Arabic languages.}}
     \label{fig:Lang}
\end{subfigure}
\vspace{-0.3cm}
\begin{subfigure}
    \centering
    \begin{minipage}{0.4\textwidth}
        \centering
        \includegraphics[width=1.1\textwidth]{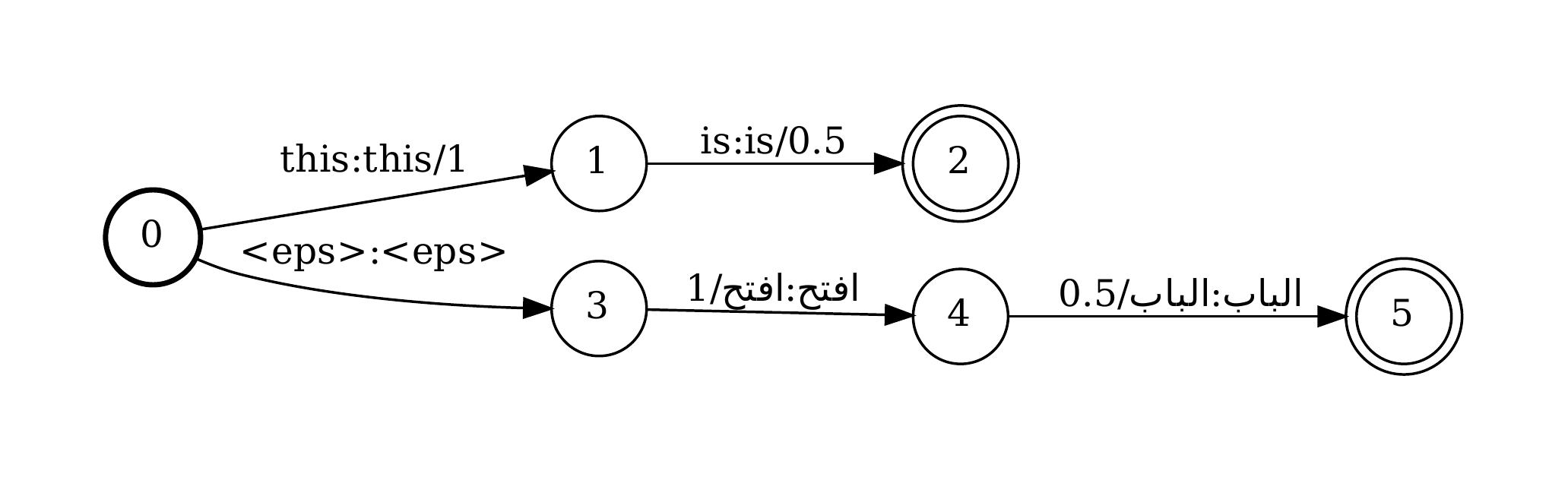}
        %\caption{Phase 1 results.}
        \label{ch:phase1_cm}
        {\small\textit{(a) Generating the union of the two graphs to allow the search in the parallel graphs.}}
    \end{minipage}\hfill
    \begin{minipage}{0.4\textwidth}
        \centering
        \includegraphics[width=1.0\textwidth]{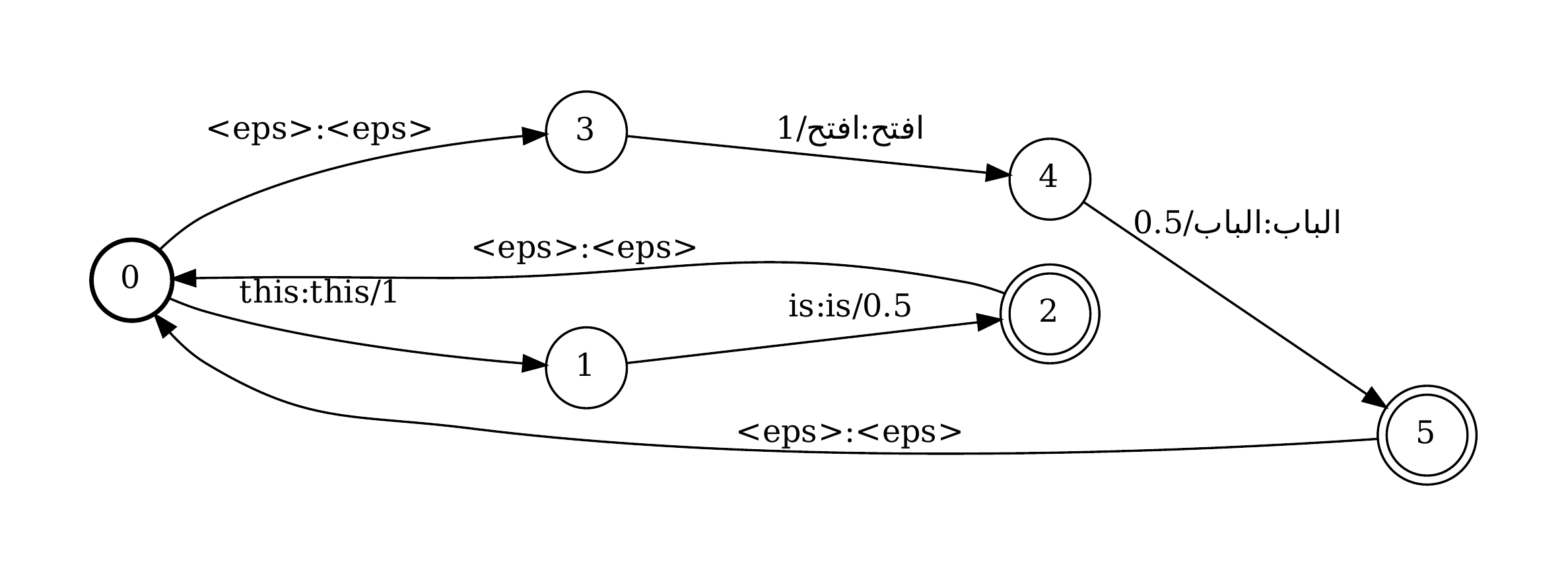}
        % \caption{Phase 2 results.}
        \label{ch:phase2_cm}
        {\small\textit{(b) Connecting final states to the initial states  via a closure operation to allow switching between languages during the search.}}
    \end{minipage}
    \vspace{-0.3cm}
    \caption{\textit{Multilingual graph building using  union and closure operations.}}
    \label{fig:mgraph}
\end{subfigure}
\end{figure*}

\begin{figure}[h]
\centering
 \includegraphics[width=0.45\textwidth]{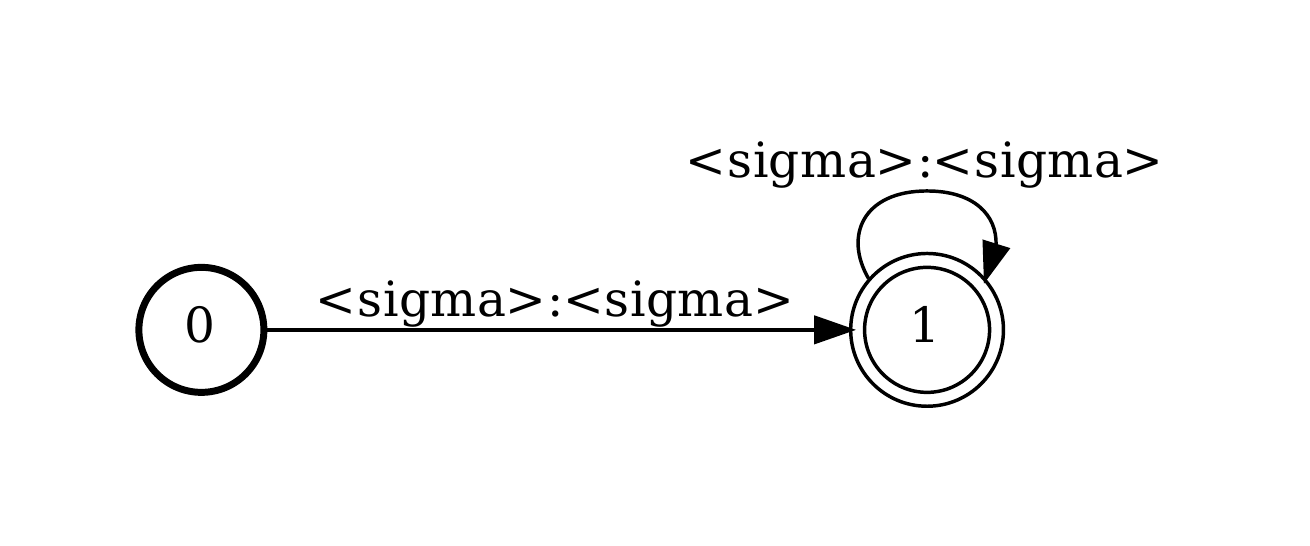} 
 \vspace{-0.8cm}
  \caption{Sigma special FST type.}\label{fig:sigma}
  \vspace{-0.3cm}
\end{figure}

For example, assume we have simple $HCLG$ graphs for English and Arabic languages as shown in Figure \ref{fig:Lang}. Then each of these graphs is composed with sigma matcher FST (shown in Figure \ref{fig:sigma}) to get rid of the paths which only have epsilons on (i.e. no symbol). This is necessary to avoid errors during decoding in the Kaldi framework. The two languages can be searched in parallel using a union operation. Finally, a closure operation would allow the transition between languages during the search process. The two operations are shown in Figure \ref{fig:mgraph}.

Allowing the transitions between languages during the search means there is no need for explicit code-switching data to train the language model (main advantage of this approach). This is a desired property of the developed multilingual system.

% \vspace{-1.8cm}

\section{Code-Switching Phenomena} \label{sec:cs_ld}

%With globalisation, code-switching in spoken conversation is becoming a common phenomena in multi-culture society. 
The two most common types of code-switching are: (\textit{i})
intersentential (switching between-utterances): the alternation in a single discourse between two languages, where the switching occurs after a sentence in the first language has been completed and the next sentence starts with a new language; and (\textit{ii}) intrasentential (within utterances): the alternation in a single discourse between two languages, where the switching occurs within the same sentence. 

These phenomena in spontaneous speech are highly unpredictable and difficult to model for both NLP and speech modules \cite{chowdhury2020effects}, thus are of great interest to the research community.
%This has been investigated in many language pairs; English-Mandarin, Indian-English, and Frisian-Dutch, as described in Section \ref{sec:intro}. However, very few studies explored both the intersentential and the intrasentential code-switching in spontaneous speech. 
One of the main reasons for such a gap in Arabic is the scarcity of both publicly available linguistic and speech resources, or lack of recipe to build models and test code-switching scenarios between Arabic and English.
%Therefore, in this study, we design two corpora -- for studying (i) inter- and (ii) intra-sentential code-switching, as discussed in details below. This is also the first attempt to study multi dialectal Arabic-English code-switching in speech recognition in hybrid and end-to-end ASR systems.

% To study the effect of such phenomena, we There is no public standard corpus or recipe to build models and test code-switching scenarios between Arabic and English. To overcome this problem, we design two corpora: 

% \vspace{-0.3cm}
\subsection{Intersentential Code-Switching Corpus}
%\noindent \textbf{Intersentential code-switching corpus}:  
We concatenate the audio files of Arabic and English%. During the concatenation
, we did not add any extra silence between sentences. It is worth mentioning that during the concatenation of the audio files, it was not possible to maintain same speakers during the transition between languages. %Hence, our code-switching corpus is harder than usual, since we have to deal with language and speaker transitions in the generated simulation audio files. 
Since the number of utterances in the English TED-LIUM3 dev/test is lower than the corresponding number in the MGB-2 dev/test as shown in Table \ref{tab:am_data}, we randomly picked sentences from the MGB-2 to match the TED-LIUM3. Despite the fact that this can be seen as artificial setup, there are many practical scenarios where non-native speakers are interviewed to comment on news, such as multinational events in Broadcast news domain \cite{ali2016mgb}.% or discuss sports, education events, such as across national borders multi-genre domain. % \cite{mgb5}. %This is happening  broadly defined as the integration of economic activity across national borders -- is widely perceived to be at a crossroad. We found this scenario happens often in broadcast news 
The \textit{EAE corpus} refers to concatenating English audio from  TED-LIUM3 with Arabic Audio from MGB-2 and English Audio from TED-LIUM3. This is to simulate $EN \longrightarrow AR \longrightarrow EN$ intersentential code-switching scenario. Same for \textit{AEA corpus}. %refers to concatenating Arabic audio from  MGB-2 with English Audio from TED-LIUM3 and Arabic Audio from MGB-2. This is to simulate Arabic$->$English$->$Arabic intersentential code-switching.
More details of the EAE and AEA\footnote{Available from \url{https://arabicspeech.org/tedl\_mgb2\_cs\_augmented/}} are present in Table \ref{tab:cs_data}.

\begin{table}[!htb]
\centering 
\scalebox{0.80}{
\begin{tabular}{c|c|c} 
Corpus ID & Total Hours & \#segments\\ 
\hline
 Dev $\mathcal{EAE}$ & 4.5h & 506 \\
 Dev $\mathcal{AEA}$ & 4.1h & 506 \\
 
 Test $\mathcal{EAE}$ & 7.8h & 1,154 \\
 Test $\mathcal{AEA}$ & 8.0h & 1,154 \\\hline
 ESCWA&2.8h&845\\
\hline
 \end{tabular}
 }
 \caption{\textit{Dev and test data used for code-switching evaluation.}}
\label{tab:cs_data}
\vspace{-0.8cm}
\end{table}

\subsection{Intrasentential Code-Switching: ESCWA Corpus}
%\noindent \textbf{Intrasentential code-switching corpus}: 
%The \textbf{ESCWA corpus} is collected in two days meetings in the United Nations Economic and Social Commission for West Asia (ESCWA) in 2019. Most of the Arabic speech is dialectal.  In addition, most of the sentences are intrasentential code-switching, where the alternation between Arabic and English is happening within the same sentence.  See Table \ref{tab:cs_data} for more details about our code-switching datasets. 2.8 hours has code-switching out of 8 hours recordings; this is an average 35\% of the meetings has intrasentential between Arabic and foreign language; this 

We studied an eight-hour corpus collected over two days of meetings of the United Nations Economic and Social Commission for West Asia (ESCWA) in 2019. 
From the data, we observed that more than 2.8 hours\footnote{Available from \url{https://arabicspeech.org/escwa/}} \cite{chowdhury2021onemodel} (in Table \ref{tab:cs_data}) of the collected speech demonstrate intrasentential code-switching, where the alternation between Arabic and English\footnote{In few cases, as with Algerian, Tunisian, and Moroccan native-speakers, the switch is between Arabic and French.} is happening within the same sentence. %Our initial analysis showed that such phenomena is present in $\approx$35\% of the dialectal Arabic speech. %Further investigation showed that on average 22\% of the segments is English content (with a total English vocabulary, $|V|$ size of 5K) and other 78\% is dialectal Arabic (Arabic $|V|$=17K words). 
We use \textit{Code-Mixing Index} (CMI) to report the level of code mixing in the ESCWA corpus.
%We measure the level of code mixing in the ESCWA corpus using both the utterance and corpus level \textit{Code-Mixing Index} (CMI) \cite{gamback2016comparing}. %This measure evaluates the complexity\footnote{Higher CMI means more complex CS.} of the mixing based on the frequency of switching and the number of different languages used in the segment. We calculated utterance level CMI ($C_u$) using Equation \ref{eq:cmi}, adapted from \cite{chowdhury2020effects}. 
%\begin{equation}
%\label{eq:cmi}
%%\vspace{-0.2cm}
%\small{
%C_u(x)=\frac{\frac{1}{2}* (N(x)-\max_{L_i \in \textbf{L} %}\left \{ t_{L_i} \right \}\left ( x \right ) ) + %\frac{1}{2}CA(x)}{N(x)}
%}
%\end{equation}
%\noindent where $N$ is the  number of tokens in utterance $x$. $L_i \in \textbf{L}$, the set of all language labels in the dataset; $\max\left \{ t_{L_i} \right \}$ represent the maximum token in the majority label class, with $1 \leq \max\left \{ t_{L_i} \right \} \leq N$; and $CA$ is the number of code alternation points in $x$; $0 \leq CA <  N$.

%We utilized the CMI to access the level of code-switching present in the ESCWA dataset. 
We  reported the corpus level CMI by simply averaging the utterance level switching.\footnote{Ignoring the switches between the utterances.} 
The details of the level of code-switching in ESCWA corpus are presented in Table \ref{tab:ESCWA-CMI}, and an example of segment in 45-100\% range is presented in Figure \ref{fig:escwa_example}.
% Furthermore, we noticed, Arabic is the dominant language in 88\% of these among 845 utterances. 
% <Ask AA, if CMI based WER is needed? -->

\begin{table}[!ht]
\centering
\scalebox{0.80}{
\begin{tabular}{l|cccc}
\hline
\multicolumn{5}{c}{ Utterance Level CMI }\\ \hline
\multicolumn{1}{c|}{CMI Range} & CA & word/Utt. & Avg. dur & \#. \\ 
\hline\hline
% 0\% & 0 &  &  &  \\
0-15\% & 2.6 & 26.8 & 12.4 & 485 \\
15-30\% & 4.7 & 24.2 & 11.4 & 289 \\
30-45\% & 7.4 & 23.2 & 10.2 & 68 \\
45-100\% & 7.7 & 14.7 & 8.6 & 3 \\ \hline
\hline
Corpus-CMI: &\multicolumn{4}{c}{ 28}\\ \hline
\end{tabular}
}
\caption{\textit{\small{Details of code-switching level of ESCWA data using CMI range. word/Utt. represents the average word count per utterance, CA is the mean number of code alternation points in utterances, \#. presents the number of utterances and and Avg. dur is average duration in seconds that belong to that particular CMI range.}}}
\label{tab:ESCWA-CMI}.
\vspace{-0.8cm}
\end{table}

\begin{table}[t]
\centering 
\scalebox{0.8}{
\begin{tabular}{c|c|c|c} 
Type & Monolingual & Multilingual& Transformer \\ 
\hline
 AR Dev & 17.2 & 17.1 & 15.49 \\
 AR Test &  16.3 & 17.5& 17.61 \\
\hline
 EN Dev & 9.2 & 9.9 &9.68\\
 EN Test &  9.1 & 9.8&8.29 \\
\hline
 \end{tabular}
 }
 \caption{\textit{WER in \% for  hybrid ASR (monolingual, multilingual) and end-to-end transformer systems.  }}
\label{tab:mon_multi_comparison}
\vspace{-0.6cm}
\end{table}

% In an eight-hour corpus collected over two days of meetings of the United Nations Economic and Social Commission for West Asia (ESCWA) in 2019, it is observed that more than 2.8 hours (in Table \ref{tab:cs_data}) of the collected speech demonstrate intrasentential code-switching, where the alternation between Arabic and English is happening within the same sentence. In few cases, as with Algerian, Tunisian, and Moroccan native-speakers, the switch is between Arabic and French. The 845 out of $\approx$2.5K sentences in ESCWA corpus indicate that intrasentential code-switching can happen in about 35\% of the dialectal Arabic speech. 

%Out of 8 hours meeting recordings, 35\% of the content -2.8 hours has code-switching. The 
%3.6 hours Arabic data, 1.6 hours English data and 2.8 code-switching data 

% The \textbf{ESCWA corpus} has 5K English words and $17$K Arabic words, with $22$-$78$\% Arabic to English words. 

\begin{table} [h]
\renewcommand\cellalign{lc}

 \centerline{
 \scalebox{0.75}{
\begin{tabular}{l|c|c}
Decoding scenario &  Dev WER & Test WER \\
\hline
%\makecell{ted3\_mgb2\_ted3 code-switching data\\+ Global G} & 20.18 & 23.27\\
%\makecell{$\mathcal{EAE}$: Global $G$} & 20.18 & 23.27\\
$\mathcal{EAE}$: Global $G$ WFST & 20.2 & 23.3\\
%\makecell{$\mathcal{EAE}$: Multi-graph  } & 19.93 &  22.97\\
$\mathcal{EAE}$: Multi-graph  WFST & 19.9 &  23.0\\
$\mathcal{EAE}$: Transformer  & 52.9 & 51.1 \\
$\mathcal{EAE}$: Oracle language diarization & 12.9  & 13.5  \\
\hline
%\makecell{$\mathcal{AEA}$: Global $G$} & 24.15 & 28.99\\
$\mathcal{AEA}$: Global $G$ WFST & 24.2 & 29.0\\
%\makecell{$\mathcal{AEA}$: Multi-graph  } & 23.82 &  28.69\\
$\mathcal{AEA}$: Multi-graph  WFST & 23.8 &  28.7\\
$\mathcal{AEA}$: Transformer  & 54.2 & 54.5 \\
$\mathcal{AEA}$: Oracle language diarization & 16.0  & 17.2  \\
\hline
ESCWA  Global $G$ WFST&\multicolumn{2}{c}{63.2}\\
ESCWA  Multi-graph WFS&\multicolumn{2}{c}{65.2}\\
\textbf{ESCWA transformer} & \multicolumn{2}{c}{\textbf{50.1}}\\
\hline
\end{tabular}}
}
\caption{Decoding performance for intersentential and intrasentential code-switching tasks. WER in \%.} \label{tb:CORRECTION_RESULTS} 
 \vspace{-0.8cm}
\end{table}

\section{Experiments and Results} \label{sec:exp}

Table \ref{tab:mon_multi_comparison} shows the WER of the monolingual, the multilingual, and the transformer systems. The decoding in the multilingual system is based on  the Global G approach, where the language model graph is built from a multilingual text. It is noticeable that the transformer system WER results are slightly better than the hybrid system.%are close to  the other systems.  

%The intersentential code-switching was tested using the generated datasets from the  TED-LIUM3 and the  MGB-2 data. The  intrasentential code-switching was tested using the ESCWA test data.

We tested four methods to evaluate our multilingual decoding strategies: (\textit{i}) the Global $G$ approach, where  the language model is trained on a multilingual corpus; (\textit{ii}) the multi-graph approach, where we build $HCLG$ graph for each language, combined using union operation;% then, by using a union operation, it is possible to search the languages in parallel. In addition, we allow the transition between languages during the search using a Kleene closure operation on the generated multi-graph from the union operation
(\textit{iii}) the end-to-end transformer architecture;% , where sequence to sequence system is used for decoding
and (\textit{iv}) finally, assuming that we have the oracle language  code-switching, we build mono-language models for Arabic and English and decoded the split dev/test sets. This setup provides an upper bound on the code-switching results reported using the global $G$ and multi-graph approaches.   

Results are shown in Table \ref{tb:CORRECTION_RESULTS}. For  intersentential code-switching, the  WFST based decoding strategies are doing much better than the transformer system. This can be explained as the attention mechanism has never encountered a long Arabic sentence followed a long English sentence in the training data set. In addition, the multi-graph approach led to the best results. Hence, the WFST decoding strategies may be more suitable for  the intersentential code-switching tasks. On the other hand, the transformer system results for  intrasentential code-switching task evaluated using the ESCWA dataset are better than the WFST decoding strategies. The transformer system can respond quickly to the rapid change in the language during talking. This is not the case for WFST decoding systems, where it is difficult to switch the language model quickly during the decoding. In general, high WER for the ESCWA dataset is expected since there is acoustic conditions mismatch between the train and test datasets. Also, the data is dominated by dialectal Arabic with intrasentential code switching. Such intrasentential phenomena can affect the performance of ASR more than intersentential code-switching, as clearly observed from the ASR performance.
% \newline

\begin{figure}[h]
\centering
\vspace{-0.2cm}
%   \scalebox{0.5}{
 \includegraphics[width=\linewidth]{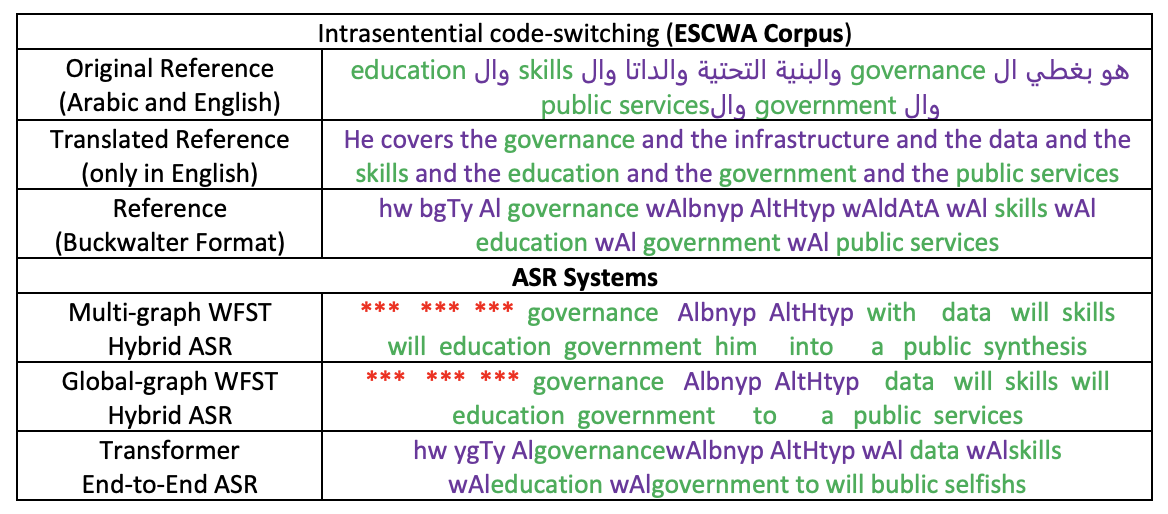}
% \scalebox{0.5}{\includegraphics{ESCWA_CS_Example_ASR_RESULTS.pdf}}
 
 \vspace{-0.4cm}
  \caption{Example of intrasentential code-switching from ESCWA Corpus. Text is in Arabic, English and Buckwalter. %Buckwalter is a one-to-one mapping allowing non Arabic speakers to understand Arabic scripts, and it is also left-to-right, making it easy to render on most devices.
  }\label{fig:escwa_example}
  \vspace{-0.2cm}
\end{figure}

%\subsection{Intersentential Example}
\noindent \textbf{Intrasentential Example:} 
%For deeper undertanding, 
Figure \ref{fig:escwa_example} presents an example from the ESCWA corpus with dialectal Arabic along with English. The 16-words sentence has a duration of 8.5 seconds with $9$ switching points. In fact, it can be argued that there are $10$ switching points, since the word data is written in Arabic script, while it is an English word%\footnote{Given that dialectal Arabic does not have a clearly defined orthography, different people tend to write some English words in Arabic script. Therefore, instead of developing strict guidelines to ensure a standardized orthography, we allow for freedom in choosing Arabic or English script.}.
The hybrid speech-to-text system in the global graph approach and the multi-graph approach are not able to get more than two switching points, while the end-to-end transformer is able to detect nine switching points. %In terms of performance, as shown in 
Table \ref{tab:example_WER} shows the global-graph WFST hybrid system is achieving the lowest WER 62.5\%, while the end-to-end achieves 81.2\%. However, by looking at the character error rate (CER) and code-switching points, it is clear that the end-to-end is outperforming both of the hybrid systems considerably. This can be addressed using transliterated WER. 
%addressed in evaluation by using a multi-reference WER estimation (MR-WER) \cite{ali2015multi}. %Refer to appendix 1 for further detailed  analysis.

%This can also be happening due to the lack of standardization in dialectal Arabic orthography, especially when it is combined with English

\begin{table}[!htb]
\centering 
\scalebox{0.8}{
\begin{tabular}{c|c|c} 
System & WER & CER \\ 
\hline
 Multi-graph WFST & 81.3 & 40.0\\
 Global-graph WFST & 62.5 & 26.7 \\
 Transformer & 81.1 & 13.3 \\
 \hline
 \end{tabular}}
 \caption{\textit{WER and CER in \% for the intrasentential code-switching example in Figure \ref{fig:escwa_example} }}
\label{tab:example_WER}
\vspace{-0.3cm}
\end{table}

% \newline

\section{Conclusions}\label{sec:CONCLUSION}
In this paper, we %developed
proposed an innovative new strategy for multilingual ASR speech recognition. In particular, we implemented three strategies to recognize code-switching speech. The first strategy is to decode the speech using  a global language model built from a multilingual text. This approach serves as our baseline system. Our innovative framework deploys a multi-graph  approach  in  the  weighted  finite  state  transducers (WFST) framework.  Using a closure operation, our decoder can switch between languages during the decoding process, and the output of this process is a bilingual/multilingual text that depends on the spoken audio. The third strategy is to decode the speech using a  powerful transformer system. Given a code-switching scenario between Arabic and English languages, the WFST decoding approaches were more suitable for the intersentential code-switching datasets. Moreover, the transformer system was doing well for the intrasentential code-switching task. The transformer system seems to very slow in training and decoding.

\bibliographystyle{IEEEtran}

\bibliography{mybib}

\end{document}